\documentclass[11pt, a4paper, logo, copyright, nonumbering]{lti}
\usepackage[authoryear, sort&compress, round]{natbib}
\usepackage{dblfloatfix}
\usepackage{ulem}
\usepackage{caption}
\usepackage{dramatist}
\usepackage{xspace}
\usepackage{pifont} 
\usepackage{multirow}
\usepackage{tcolorbox}
\usepackage{xltabular}
\usepackage{longtable}
\usepackage{hyperref}
\usepackage{tikz}
\usetikzlibrary{positioning}
\usepackage{amsfonts}
\usepackage{amsmath}
\usepackage{amssymb}
\usepackage{lineno}
\usepackage{multirow}
\usepackage{adjustbox}

\usepackage[bottom]{footmisc}

\usepackage{CJKutf8}
\usepackage{subfigure}
\usepackage{setspace}

\usepackage{dsfont}
\usepackage{array}
\usepackage{tabularx}
\usepackage{subfigure}
\usepackage{xcolor}
\usepackage{wrapfig}
\usepackage{booktabs}
\newtheorem{definition}{Definition}

\definecolor{ltired}{HTML}{780000}

\usepackage{lipsum}
\usepackage{listings}
\usepackage{fvextra}
\usepackage{multicol}
\usepackage{makecell}
\usepackage{forest}
\usepackage{xcolor}
\usepackage{dirtree}

\makeatletter
\def\@BTrule[#1]{%
  \ifx\longtable\undefined
    \let\@BTswitch\@BTnormal
  \else\ifx\hline\LT@hline
    \nobreak
    \let\@BTswitch\@BLTrule
  \else
     \let\@BTswitch\@BTnormal
  \fi\fi
  \global\@thisrulewidth=#1\relax
  \ifnum\@thisruleclass=\tw@\vskip\@aboverulesep\else
  \ifnum\@lastruleclass=\z@\vskip\@aboverulesep\else
  \ifnum\@lastruleclass=\@ne\vskip\doublerulesep\fi\fi\fi
  \@BTswitch}
\makeatother

\addto\extrasenglish{
}

 {\begin{list}{}%
         {\setlength{\leftmargin}{#1}}%
         \item[]%
 }
 {\end{list}}
 
\newcommand{\cmu}{\raisebox{-2.5mm}{\includegraphics[width=3.5mm]{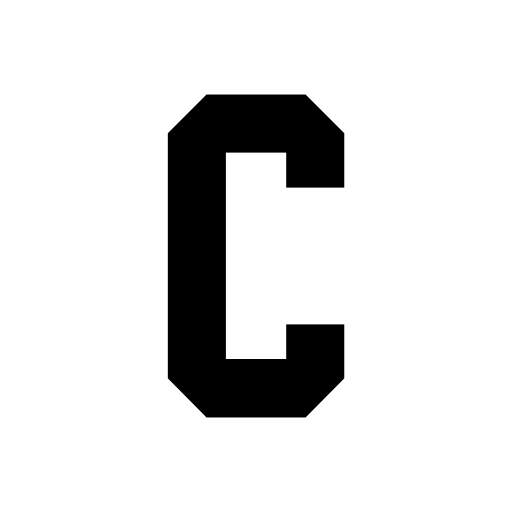}}}
\newcommand{\stfd}{\raisebox{-0.5mm}{\includegraphics[width=2.3mm]{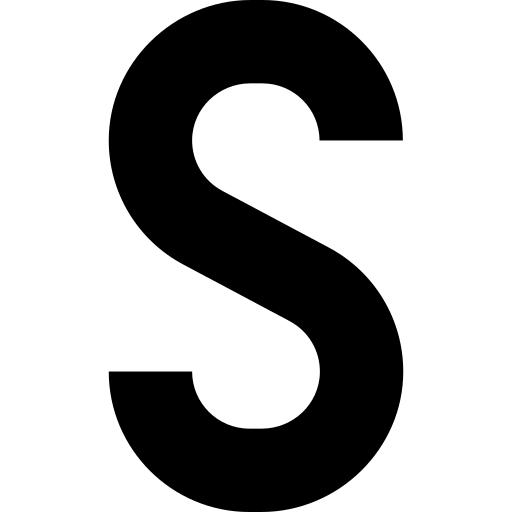}}}

\usepackage{color, colortbl}
\usepackage{color-edits}
\addauthor{df}{cyan}
\addauthor{gn}{magenta}
\addauthor{dy}{blue}
\addauthor{zw}{orange}
\addauthor{vc}{red}

\title{\centering How Well Does Agent Development Reflect Real-World Work?}

\author{\centering
\normalsize{\textbf{Zora Z. Wang$^{\cmu}$ \quad Sanidhya Vijayvargiya$^{\cmu}$ \quad Aspen Chen$^{\cmu}$\\
{Hanmo Zhang}$^{\cmu}$ \quad {Venu Arvind Arangarajan}$^{\cmu}$  \quad Jett Chen$^{\cmu}$ \\
{Valerie Chen}$^{\cmu}$ \quad {Diyi Yang}$^{\stfd}$ \quad {Daniel Fried}$^{\cmu}$ \quad {Graham Neubig}$^{\cmu}$}} 
\\
\vspace{0.3em}
$^{\cmu}$Carnegie Mellon University \quad $^{\stfd}~$Stanford University
\\
\vspace{0.3em}
\small{\texttt{zhiruow@cs.cmu.edu}} \\
\url{https://zorazrw.github.io/ai4work/}
}

\renewcommand{\phi}{\varphi}

\renewcommand{\epsilon}{\varepsilon}
\renewcommand{\imath}{\mathrm{i}}

\newlength{\restsubwidth}
\newlength{\restsubheight}
\newlength{\restsubmoreheight}
\newcommand{\rest}[2]{%
        \settowidth{\restsubwidth}{\ensuremath{#2}}
        \settoheight{\restsubheight}{\ensuremath{{}_{#2}}}
        \ensuremath{{#1\hskip 0.5pt}_{\vrule\kern2pt\parbox[b][%
        4pt][b]{\the\restsubwidth}{%
                        \ensuremath{{}_{#2}}}}}
        }

\begin{abstract}
AI agents are increasingly developed and evaluated on benchmarks relevant to human work, yet it remains unclear how representative these benchmarking efforts are of the labor market as a whole. In this work, we systematically study the relationship between agent development efforts and the distribution of real-world human work by mapping benchmark instances to work domains and skills. 
We first analyze 43 benchmarks and 72,342 tasks, measuring their alignment with human employment and capital allocation across all 1,016 real-world occupations in the U.S. labor market. 
We reveal substantial mismatches between agent development that tends to be programming-centric, and the categories in which human labor and economic value are concentrated.
Within work areas that agents currently target, we further characterize current agent utility by measuring their autonomy levels, providing practical guidance for agent interaction strategies across work scenarios.
Building on these findings, we propose three measurable principles for designing benchmarks that better capture socially important and technically challenging forms of work: coverage, realism, and granular evaluation.\footnote{text}
\end{abstract}

\begin{document}

\maketitle

\section{Introduction}

AI agents have made rapid progress in their accuracy and autonomy on tasks such as web navigation \citep{deng2023mindweb,zhou2024webarena} and versatile computer use \citep{xie2024osworld}. A central motivation behind the development of such agents is to enhance human productivity across diverse forms of work \citep{wang2025ai}, such as software engineering \citep{deng2025swe} and knowledge-intensive research \citep{wei2025browsecomp}.
As agent capabilities advance, benchmark design has primarily evolved by increasing task complexity \citep{miyai2025webchorearena}, extending the amount of work done in each task \citep{yoran2024assistantbench}, or introducing more challenging environments \citep{xu2024theagentcompany}. 
While these changes aim to better stress-test agents, it remains unclear whether this scaling of difficulty meaningfully reflects the structure and demands of human work \citep{shao2025future}, or how performance on existing benchmarks translates into practical relevance for real-world jobs. 
Even when specific benchmarks explicitly target work-related activities \citep{xu2024theagentcompany,patwardhan2025gdpval}, they do not rely on a standard method for categorization of work, hindering cross-benchmarks comparisons, and obscuring which work domains (e.g., administrative support) and underlying skills (e.g., information gathering) are being approximated, and who ultimately benefits from improved agent performance.

To address this gap, in this paper, we present a systematic framework that situates agent benchmarks within the broader landscape of human work, particularly in the context of the U.S. labor market. Specifically, we map individual benchmark examples to domains and skills using occupational taxonomies derived from the O*NET database \citep{onet2024}, a U.S. government resource that catalogs real-world work activities at multiple levels of granularity. 
Because benchmark tasks often specify the task but not the activities required to actually complete it, this mapping requires disentangling composite activities and aligning informal descriptions with structured occupational labels. We implement this via large language model (LLM)-based annotation with manual quality verification, enabling scalable processing across benchmarks (\S\ref{sec:2:bridge-agent-work}).

Building on this mapping framework, we conduct a large-scale analysis spanning 43 agent benchmarks, 72,342 task instances, and 1,016 real-world occupations, providing a holistic characterization of how current agent development efforts align with the broader landscape of the U.S. labor market.
We find that existing agent benchmarks are heavily concentrated in the computer and mathematical domain, which accounts for only 7.6\% of U.S. employment, while other highly digitized and economically significant areas, such as management and legal work, remain substantially underrepresented. These domains are not only central to organizational decision-making and economic coordination, but also pose distinct technical challenges, such as ambiguous objectives and verification with long-horizon dependencies.
We observe a similar pattern at the skill level: agent development disproportionately targets a small set of fine-grained skills that together account for less than 5\% of total U.S. employment, whereas widely prevalent skills such as interpersonal interaction, which permeate most jobs, are largely absent (\S\ref{sec:4:skewed-dev}).

Even within the work areas that current benchmarks target, tasks vary substantially in contextual and procedural complexity, making it hard to translate performance metrics into actionable insights of agent capability boundaries.
We thus propose a unified task complexity measure that places agent activities on a common scale. This enables measurement of agent autonomy, defined as the performance frontier across tasks of increasing complexity.
We analyze agent trajectories across benchmarks to depict their autonomy levels by domain, skills, agent framework, and backbone LMs, providing actionable guidance for selecting the appropriate autonomy level to interact with them (\S\ref{sec:3:autonomy}).

Building on these findings, we propose three measurable benchmark design principles---domain and skill coverage, task realism and complexity, and granular evaluation---to better capture the breadth and structure of real-world work (\S\ref{sec:5:discussion}). 
Together, these contributions establish a systematic framework for evaluating how well agent benchmarks reflect real-world work, and provide actionable tools for guiding both benchmark design and agent development toward more representative and socially grounded progress.

\section{Bridging Agents to Real-World Work}
\label{sec:2:bridge-agent-work}

To translate agent development to real-world utility, we depict the landscape of human work (\S\ref{sec:2.1:work-landscape}) and situate agent building efforts in the same space for comparison (\S\ref{sec:2.2:agent-effort}).

\subsection{The Landscape of Human Work}
\label{sec:2.1:work-landscape}

\begin{figure}[t!]
\centering
    \includegraphics[width=0.7\textwidth]{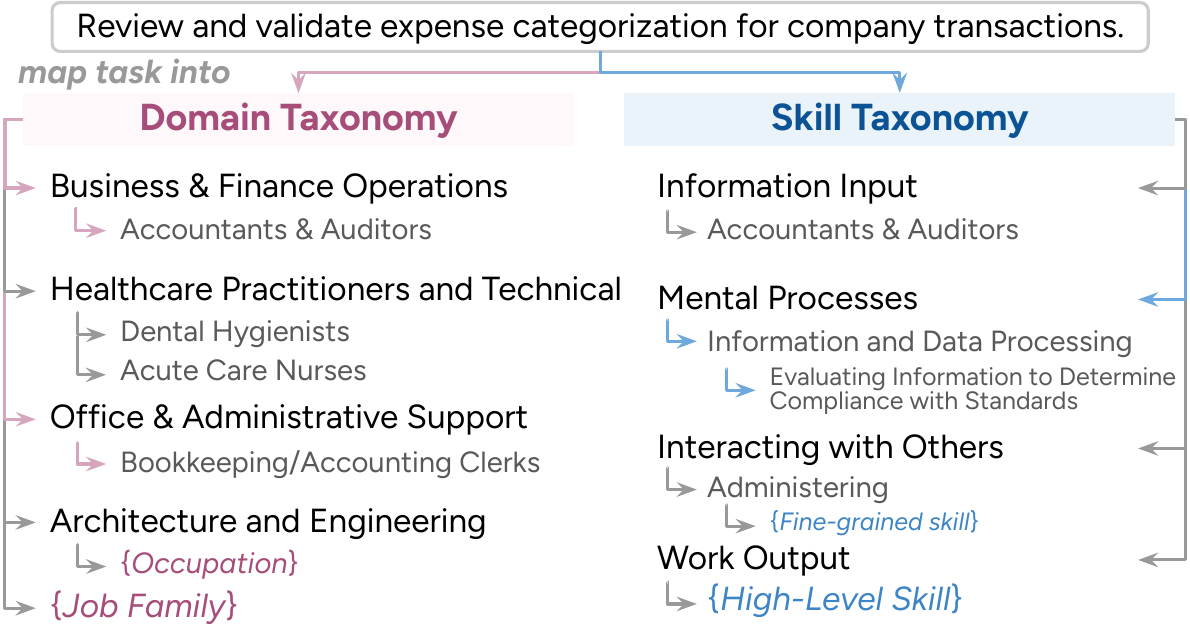}
    \vspace{-2mm}
\caption{Mapping agent benchmarks to work domains and skills in O*NET.}
\label{fig:taxonomy}
\vspace{-2mm}
\end{figure}

We depict the landscape of work along two complementary dimensions, domain and skill, by constructing two taxonomies based on the O*NET database \citep{onet2024}, as in \autoref{fig:taxonomy}. Crucially, to ensure a faithful representation of real-world work, these taxonomies are defined purely from human occupations and not biased by existing agent development efforts.

\noindent \textbf{Domain-Based Taxonomy} \quad
We build a domain taxonomy $T^d$ based on the job family and task requirement annotations (as in \autoref{fig:taxonomy}, left) in O*NET. Our taxonomy connects work domain variety across granularities: from high-level industries (e.g., Business and Financial Operations), pertinent occupations (e.g., Accountants, Budget Analysts), to concrete tasks performed in practice (e.g., prepare adjusting journal entries).
Overall, the domain taxonomy $T^d = (V^{(0)}, V^{(1)}, V^{(2)}, V^{(3)}, E)$ has 23 job families ($|V^{(1)}|$), of which 743 occupations ($|V^{(2)}|$) and 5806 task descriptions ($|V^{(3)}|$) involve computer use; $V^{(0)}$ denotes the ``domain'' root and $E$ is the set of tree edges.

\noindent \textbf{Skill-Based Taxonomy} \quad
In the context of agent work, we define skill as a concrete sequence of actions performed to achieve a goal. We construct the skill taxonomy $T^s$ starting from O*NET's Work Activities (WA) taxonomy\footnote{Although O*NET labels skills in a cognitive sense (\url{https://www.onetonline.org/find/descriptor/browse/2.A}), we adopt this formulation as it offers a more tangible basis for agent analysis.} organized in four general categories $V^{(1)} = \{$ \textit{information input}, \textit{interacting with others}, \textit{mental processes}, \textit{work output}$\}$, each expanding into multiple detailed activities. We further expand the taxonomy with O*NET's Detailed Work Activities. 
The resulting taxonomy contains three layers of skills of progressively finer granularity, with 4, 9, and 41 nodes, as summarized in \autoref{fig:taxonomy} (right).

For both taxonomies, we denote a path to be a sequence of categories from the root to a granular leaf one: $p = (v_0, v_1, \cdots, v_l), v_0 = r, (v_i, v_{i+1}) \in E$, where $l$ represents the total number of layers in the taxonomy tree. We explicitly preserve the full path to model the coarse-to-fine structure of work, enabling analysis at various levels of abstraction. 
By cross-indexing tasks to these two taxonomies, we can obtain complementary views to associate them with work: domains reflect job-specific contexts, while skills capture processes that generalize across jobs.

\noindent \textbf{Human Employment and Capital Distribution} \quad
Human employment and economic value are not equally distributed across domains and skills. To capture this variance, we collect employment and capital statistics for each category in the domain and skill taxonomies.
For the domain taxonomy, since each leaf category corresponds to an occupation, we obtain occupation employment and median salary data  (latest update in 2024) from the U.S. Bureau of Labor Statistics (BLS) \citep{laborstats2024}. Because both BLS and O*NET use the Standard Occupational Classification (SOC) system, we directly align each leaf occupation in our domain taxonomy with its corresponding employment and wage statistics via shared SOC codes.
We aggregate these counts upward to estimate employment and capital at broader domain levels.
For the skill taxonomy, we compute employment and capital estimates for each fine-grained work activity by weighting occupational data using the corresponding activity-level importance scores provided by O*NET.\footnote{E.g., \url{https://www.onetonline.org/find/descriptor/result/4.A.1.b.3}} We then aggregate these weighted values to higher-level skill categories. Importantly, these estimates do not represent the exact number of workers performing a given skill in their daily work; rather, they approximate the relative importance of each skill across the overall labor market.

To provide a holistic view of real-world work, we retain domains and skills involving physical labor (e.g., manufacturing) in both taxonomies. To explicitly model digital and physical forms of work, we separately quantify \textit{digital} and \textit{physical} labor. Concretely, for each occupation, we examine its associated task requirements and classify each task as digital or physical by prompting an LLM\footnote{We used \texttt{claude-sonnet-3.7} \citep{claude37} for efficiency.} to assign a \texttt{DIGITAL} or \texttt{PHYSICAL} label based on the occupation name and task description. We then compute, for each occupation, the proportion of digital tasks and weight this proportion by employment and capital statistics to derive aggregate measures for various levels of domains and skills.
Find the full taxonomies and calculation process in \S\ref{app:a:work-landscape}.

\subsection{Agent Development Effort for Work}
\label{sec:2.2:agent-effort}

Benchmarks are a primary driver of agent development, shaping both the tasks on which agents are trained and the settings in which they are evaluated. To characterize the current landscape of digital agent development for work-related activities, we aggregate representative agent benchmarks and map them onto the work taxonomies introduced above (\S\ref{app:b:agent-dev-details}).

We aim to comprehensively aggregate existing agent benchmarks related to human work (\autoref{tab:benchmark-overview}). To ensure consistency with the core definition of agents, we apply two minimal inclusion criteria: (i) \textit{agentic}, in that the system operates within an interactive environment (e.g., a web browser or desktop) and follows an observation-action loop during task execution; and (ii) \textit{work-related}, meaning that at least a subset of tasks corresponds to real-world work activities.

\begin{table}[t!]
\centering
\small
\resizebox{0.98\linewidth}{!}{
\begin{tabular}{clrccc}
\toprule
\multicolumn{1}{c}{\multirow{2}{*}{\bf Category}} & \multicolumn{1}{c}{\multirow{2}{*}{\bf Benchmark}} & \multicolumn{2}{c}{\bf Size} & \multicolumn{2}{c}{\bf Coverage (\%)} \\
{} & {} & \multicolumn{1}{c}{total} & {sampled} & {domain} & {skill} \\
\midrule
\multirow{8}{*}{\makecell[c]{General\\Digital\\Work}} & {TheAgentCompany \citep{xu2024theagentcompany}} & {175} & {175} & {30.4} & {56.1} \\
{} & {GDPval \citep{patwardhan2025gdpval}} & {220} & {220} & {\bf 47.8} & {\bf 58.5} \\
{} & {Remote Labor Index \citep{mazeika2025remote}} & {9} & {$~~~~$9} & {21.7} & {17.1} \\
{} & {WorkArena \citep{drouin2024workarena}} & {16,833} & {300} & {26.1} & {22.0} \\
{} & {OfficeBench \citep{wang2024officebench}} & {300} & {300} & {39.1} & {43.9} \\
{} & {CRMArena \citep{huang2025crmarena}} & {1,170} & {300} & {30.4} & {22.0} \\
{} & {EnterpriseBench \citep{vishwakarma2025can}} & {483} & {340} & {39.1} & {43.9}\\
{} & {GitTaskBench \citep{ni2025gittaskbench}} & {54} & {$~~$54} & {30.4} & {17.1} \\
\midrule
\multirow{12}{*}{\makecell[c]{Web \&\\Mobile\\Navigation}} & {OSWorld \citep{xie2024osworld}} & {418} & {390} & {30.4} & {46.3} \\
{} & {WebVoyager \citep{he2024webvoyager}} & {643} & {300} & {30.4} & {34.1} \\
{} & {WebArena \citep{zhou2024webarena}} & {812} & {300} & {39.1} & {43.9} \\
{} & {Mind2Web \citep{deng2023mindweb}} & {1,015} & {300} & {30.4} & {31.7} \\
{} & {WebShop \citep{yao2022webshop}} & {12,251} & {300} & {$~~$8.7} & {12.2} \\
{} & {VisualWebArena \citep{koh2024visualwebarena}} & {910} & {300} & {30.4} & {29.3} \\
{} & {WebLINX \citep{lv2024weblinx}} & {99} & {$~~$99} & {30.4} & {24.4} \\
{} & {AppWorld \citep{trivedi2024appworld}} & {732} & {300} & {30.4} & {36.6} \\
{} & {AssistantBench \citep{yoran2024assistantbench}} & {214} & {214} & {43.5} & {17.1} \\
{} & {SPA-Bench \citep{chen2025spabench}} & {340} & {300} & {34.8} & {29.3} \\
{} & {MMInA \citep{tian2025mmina}} & {1,050} & {300} & {26.1} & {24.4} \\
{} & {WebChoreArena \citep{miyai2025webchorearena}} & {718} & {300} & {34.8} & {22.0} \\
\midrule
\multirow{1}{*}{\makecell[l]{Information}} & {GAIA \citep{mialon2024gaia}} & {466} & {300} & {\bf 47.8} & {26.8} \\
\midrule
\multirow{2}{*}{\makecell[c]{Planning}} & {TravelPlanner \citep{xie2024travelplanner}} & {1,000} & {300} & {$~~$8.7} & {44.4} \\
{} & {DeepPlanning \citep{zhang2026deepplanning}} & {120} & {120} & {$~~$4.3} & {33.3} \\
\midrule
\multirow{8}{*}{\makecell{Software\\Engineering}} & {SWE-bench \citep{jimenez2024swebench}} & {2,294} & {300} & {13.0} & {17.1} \\
{} & {TerminalBench \citep{tbench2025}} & {232} & {232} & {39.1} & {29.3} \\
{} & {ColBench \citep{zhou2025sweet}} & {20,020} & {860} & {43.5} & {48.8} \\
{} & {SWE-Lancer \citep{miserendino2025swe}} & {198} & {198} & {$~~$8.7} & {12.2} \\
{} & {SWE-Bench MM \citep{yang2025swebench}} & {510} & {300} & {17.4} & {24.4} \\
{} & {SWE-Bench Pro \citep{deng2025swe}} & {731} & {300} & {17.4} & {29.3} \\
{} & {MLE-Bench \citep{chan2024mle}} & {82} & {$~~$82} & {34.8} & {19.5} \\
{} & {SWT-Bench \citep{mündler2024swtbench}} & {433} & {300} & {$~~$4.3} & {19.5} \\
\midrule
\multirow{8}{*}{Science} & {DiscoveryBench \citep{majumder2025discoverybench}} & {344} & {300} & {30.4} & {17.1} \\
{} & {ScienceAgentBench \citep{chen2025scienceagentbench}} & {102} & {102} & {13.0} & {14.6} \\
{} & {CORE-Bench \citep{siegel2024corebench}} & {45} & {$~~$45} & {13.0} & {4.9} \\
{} & {SciCode \citep{tian2024scicode}} & {80} & {$~~$80} & {13.0} & {9.8} \\
{} & {MLGym \citep{nathani2025mlgym}} & {19} & {$~~$19} & {$~~$8.7} & {17.1} \\
{} & {DiscoveryWorld \citep{jansen2024discoveryworld}} & {59} & {$~~$59} & {$~~$8.7} & {29.3} \\
{} & {LabBench \citep{laurent2024lab}} & {1,970} & {300} & {21.7} & {19.5} \\
{} & {SUPER \citep{bogin2024super}} & {801} & {300} & {$~~$8.7} & {12.2} \\
\midrule
\multirow{2}{*}{Social} & {Sotopia \citep{zhou2024sotopia}} & {3,860} & {300} & {39.1} & {46.3} \\
{} & {STSS \citep{wang2024towards}} & {40} & {$~~$40} & {26.1} & {34.1} \\
\midrule
\multirow{2}{*}{\makecell[c]{Physical}} & {Behavioral-1K \citep{li2023behavior}} & {$~~~~$50} & {$~~$50} & {39.1} & {22.2} \\
{} & {FieldWorkArena \citep{moteki2025fieldworkarena}} & {440} & {300} & {34.8} & {24.4} \\
\midrule
\multicolumn{2}{c}{\it Total} & {72,342} & {10,288} & {56.5} & {85.4} \\
\bottomrule
\end{tabular}
}
\caption{Agent benchmark sizes and their coverage on the real-world work. GDPval, despite being relatively small, has the highest domain and skill coverage across human work.}
\label{tab:benchmark-overview}
\vspace{-5mm}
\end{table}

\noindent \textbf{Mapping Benchmark Examples to Work} \quad
We map each benchmark example $e$ (a task definition in natural language) to one or more paths $P = P^d \cup P^s = \{p^d\} \cup \{p^s\}$ in both taxonomies, where each path traces a hierarchy from a top-level category down to the most granular domain or skill represented in the taxonomy, thereby capturing both coarse and detailed aspects of the work reflected by the task.
We perform this mapping $\phi$ using an LLM \footnote{We used \texttt{gpt-5} \citep{gpt5} for higher-precision mapping.} by providing the NL task instruction with a flattened representation of the taxonomy, and asking the model to identify all relevant taxonomy paths covered by the task, as $\phi(e, {\it LM}) \rightarrow P$.

\noindent \textbf{Calculating Coverage} \quad
We define \textit{coverage} of a set of examples $\mathcal{E} = \{e\}$ on a taxonomy as the percentage of unique paths being covered in the taxonomy tree $c(\mathcal{E}, T) = \frac{\phi(\mathcal{E})}{|\mathcal{P}(T)|}$. In other words, in the example mapping in \autoref{fig:taxonomy}, \textit{coverage} calculates the percentage of colored (i.e., mapped) paths among all paths in the taxonomy tree.
Across all benchmarks, examples collectively cover a limited 56.5\% of the domain taxonomy but a substantially broader 85.4\% of the skill taxonomy. This difference arises because domains correspond to occupation-specific contexts, whereas skills represent cross-occupational processes: a single benchmark example may activate multiple general skills (e.g., getting information, coordination, analyzing data) even if it is situated within a narrow occupational domain.
Although our taxonomy includes both digital and physical skills, we find many O*NET skill categories are modality-agnostic; thus, benchmarks focused on digital tasks still achieve broad skill coverage, even if physical realizations are underrepresented.

\noindent \textbf{Coverage-Aware Sampling} \quad
Some benchmarks contain a large number of task instances that are largely homogeneous, which would impose substantial computational cost when analyzed at scale across many benchmarks, while contributing to merely a subset of the taxonomies repetitively. To save cost while reflecting coverage similar to that of the original benchmark, we develop a coverage-guaranteed sampling strategy that continues to sample new batches of examples (of size 5) until coverage increases slower than $\Delta = 0.1$.
This strategy ensures that the sampled subset remains representative of the benchmark’s work diversity, while being more cost-efficient.
We validate the robustness of our coverage-aware sampling via permutation tests, showing stable stopping size and domain/skill coverage across resampled task orderings (\S\ref{app:b:mapping-validation}).

\begin{figure*}[t!]
\centering
    \includegraphics[width=\textwidth]{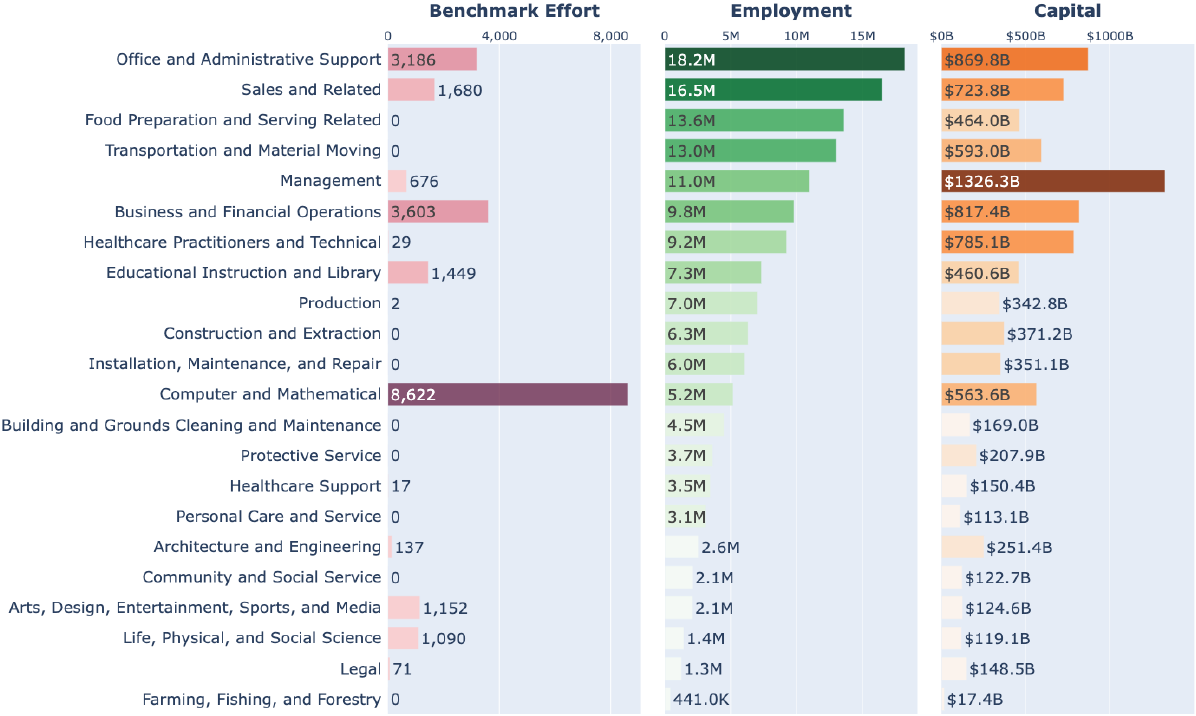}
\caption{Agent benchmarking effort is concentrated in mathematical and engineering domains, despite broad human employment and economic value across other domains.
}
\label{fig:emp-domain}
\vspace{-2mm}
\end{figure*}

\noindent \textbf{Manual Validation} \quad
We successfully map 91.2\% and 95.5\% examples into the domain and skill taxonomies; other times, the LM outputs no (8.0\% and 3.7\%) or invalid (0.8\% and 0.8\%) paths.
To examine the reliability of the LM-driven mapping process, we conduct a manual validation study by randomly sampling 90 examples across benchmarks and categorizing their mappings into four outcome types: fully correct, fully incorrect, partially correct but missing elements, or containing extraneous elements.
Two independent human annotators and the LM achieve high agreement rates of 90.9\% and 89.3\% for domain and skill mapping results, suggesting the reliability of this LM-based mapping process.
Refer to \S\ref{app:b:mapping-validation} for detailed descriptions of mapping, sampling, and verification procedures.

\section{Skewed Emphasis in Agent Development}
\label{sec:4:skewed-dev}

In this section, we analyze the alignment between agent development effort and the real-world work landscape (\S\ref{sec:4.1:domain-align}, \S\ref{sec:4.2:skill-align}).

\subsection{Benchmarks Focus on Software Engineering but Miss Other Digitized Domains}
\label{sec:4.1:domain-align}

\autoref{fig:emp-domain} compares the distribution of agent benchmark examples with real-world employment across domains. 
For each domain $d$, benchmark effort is defined as the number of benchmark examples mapped to that domain $\sum_{e \in \mathcal{E}} \mathbb{1}[{\text dom}(\phi(e, {\it LM})) = d]$, where $\phi(\cdot)$ returns a root-to-leaf taxonomy path and ${\text dom}(\cdot)$ extracts its domain-level node (\S\ref{sec:2.2:agent-effort}). 
Real-world employment $\sum_{o \in d} {\text employment}(o)$ aggregates the number of workers across occupations $o$ within domain $d$. Real-world capital is defined as $\sum_{o \in d} {\text salary}(o) \times {\text employment}(o)$ capturing the total earning-based economic value associated with occupations in domain $d$ (\S\ref{sec:2.1:work-landscape}).

Unsurprisingly for those familiar with recent agent development, agent benchmarking effort is \textit{overwhelmingly concentrated in the Computer and Mathematical domain}, featuring mainly programming tasks.
On one hand, software can be used to perform tasks across a wide variety of domains, so developing software engineering abilities has the potential to accelerate other varieties of work.
However, this concentrated focus significantly overrepresents a domain that accounts for only 7.6\% of total employment, and general-purpose software engineering benchmarks do not fully capture the domain-specific nuances in many other work areas.
As a result, large portions of the labor market remain weakly represented in current agent evaluations.

Examining domain-level digitization alongside employment reveals several domains that are heavily digitized and yet receive relatively little agent development effort.
Notably, \textit{Management, Legal, and Architecture and Engineering exhibit high ratios of digital work} (88\%, 70\%, and 71\%, respectively), \textit{but are sparsely covered} by existing benchmarks (1.4\%, 0.3\%, 0.7\% among all 19179 examples).
This gap suggests missed opportunities where agents could plausibly deliver near-term productivity gains.

Viewing work via capital distribution further exposes \textit{a disconnect between benchmarking focus and economic impact}. First, economically valuable domains, most notably Management, which also feature heavily digitized work, are underrepresented, indicating that current efforts are not targeting the highest-revenue segments of the labor market. 
Meanwhile, low-paying, labor-intensive domains such as \textit{Personal Care and Service} are likewise underexplored.

Taken together, these observations suggest that agent benchmarking effort is driven less by alignment with real-world employment structure or economic value, and more by methodological convenience.
In particular, domains with readily specified NL task instructions and easily verifiable rewards are disproportionately favored. While this focus has enabled rapid methodological progress in the areas where benchmarking is convenient, it risks skewing agent development away from domains where societal and economic impact may be greatest.

\begin{figure*}[t!]
\centering
    \includegraphics[width=\textwidth]{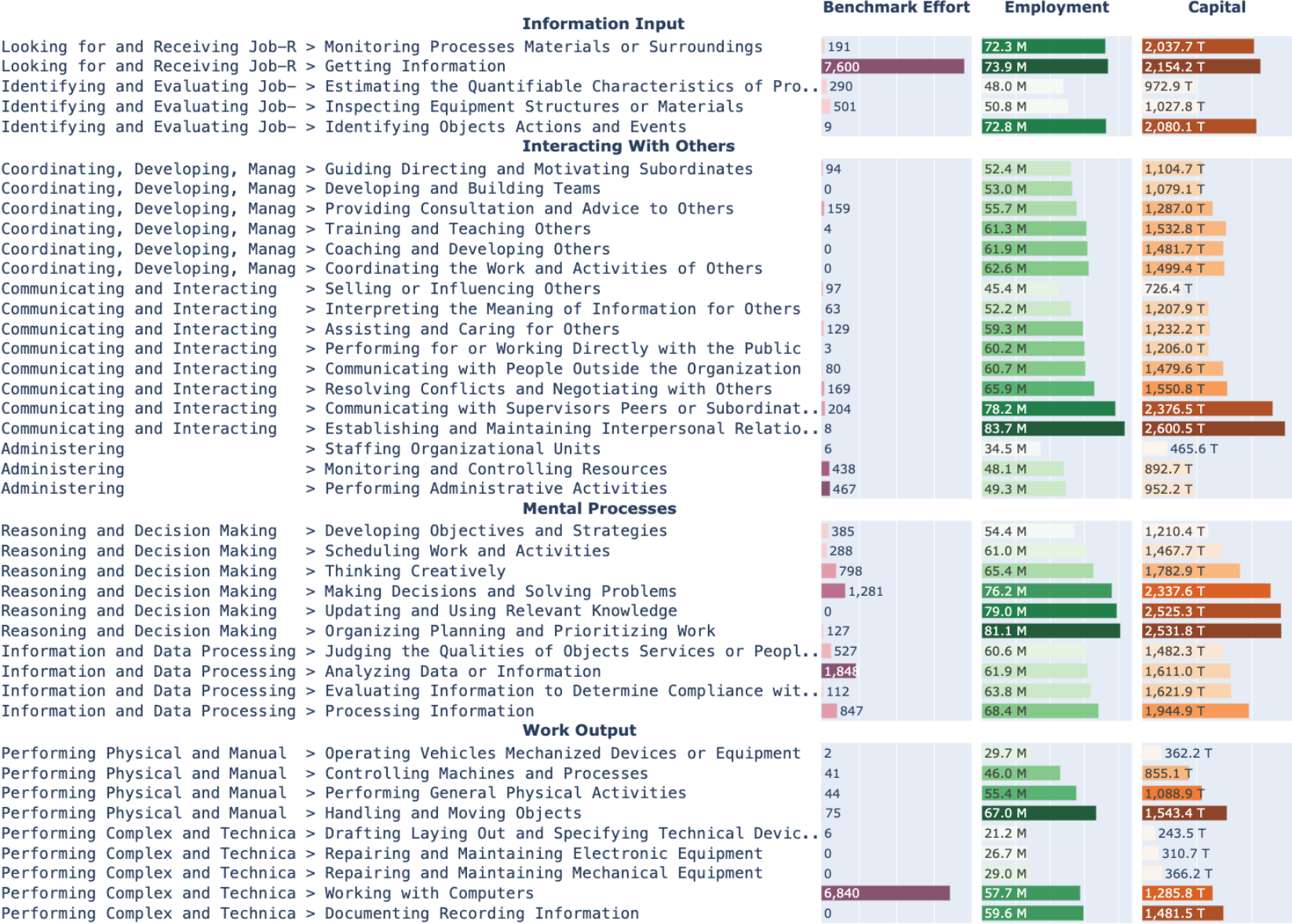}
\caption{Agent benchmarking effort emphasizes two granular skills (getting information and working with computers), which together, only cover $<5$\% human employment, creating a heavily imbalanced distribution across skills.
See \S\ref{app:2.1:onet-skills} for full skill names.
}
\label{fig:skill-leaf}
\vspace{-3mm}
\end{figure*}

\subsection{Benchmarks Overfocus on a Narrow Set of Skills}
\label{sec:4.2:skill-align}

Similarly to the domain analysis, for each skill $s$, we measure the benchmark effort for it by the number of benchmark examples mapped to that granular skill $\sum_{e \in \mathcal{E}} \mathbb{1}[{\text leaf}(\phi(e, {\it LM})) = s]$, where $\phi(\cdot)$ returns a root-to-leaf path in the skill taxonomy and ${\text leaf}(\cdot)$ extracts its leaf-level node (\S\ref{sec:2.2:agent-effort}).
Reported effective employment $\sum_{o} {\text employment}(o) \times {\text importance}(o,s)$ is the aggregation of the number of workers of each occupation $o$ weighted by the importance of skill $s$ in their job. Effective capital is defined as $\sum_{o \in s} {\text salary}(o) \times {\text employment}(o) \times {\text importance(o)}$ capturing the total earning-based economic value associated with occupations executing skill $s$ (\S\ref{sec:2.1:work-landscape}).

As shown \autoref{fig:skill-leaf}, human work typically draws on a diverse mix of skills spanning information input, mental processes, interaction with others, and work output, with no single category dominating the labor landscape. This balance reflects the multifaceted nature of real work, where tasks routinely require coordinating multiple forms of activity rather than repeatedly exercising a narrow set of capabilities.

In contrast, agent benchmarking effort is heavily concentrated on a small number of fine-grained skills. In particular, benchmark examples disproportionately target leaf-level activities such as \textit{Getting Information} within the Information Input category and \textit{Working with Computers} within Work Output, which only occupy 3.1\% and 2.4\% of employment. This concentration produces two distinct distortions. First, within otherwise broad skill categories, development effort is \textit{unevenly allocated across granular skills}, resulting in overemphasis on a few easily benchmarked activities while neglecting others at the same level of abstraction. Second, this focus \textit{crowds out entire high-level skill categories}, most notably \textit{Interacting with Others}, which receive minimal coverage despite practically pervading a wide range of real-world occupations.

\subsection{From Work Relevance to Representativeness}
\label{sec:4.4:work-relevance}

Assigning a task to a domain or skill establishes its relevance to work, but relevance alone does not imply representativeness --- whether a benchmark task captures the realistic scope, context, and complexity of the corresponding work in practice. Many benchmarks include simplified versions of work-relevant tasks that omit important contextual or procedural details.

\begin{figure*}[t!]
\centering
    \includegraphics[width=\textwidth]{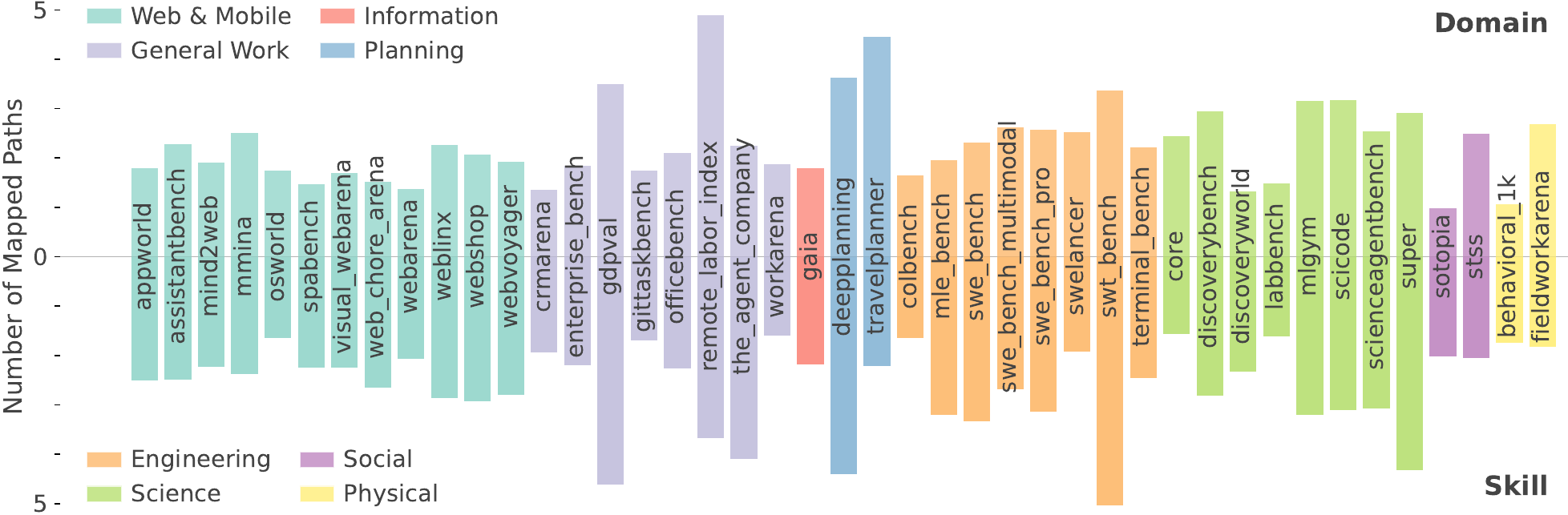}
\caption{Number of domains (top) and skills (bottom) an average benchmark example map to.}
\label{fig:no-complexity}
\vspace{-3mm}
\end{figure*}

\noindent \textbf{Domain Breath} \quad
We measure domain breadth by the number of work domains associated with each task, reflecting the breadth of contextual knowledge involved. In \autoref{fig:no-complexity} (top), although most examples (77.1\%) map to more than 1 domain, only 8.5\% of examples span more than 3 domains (e.g., \texttt{GDPval}), indicating limited cross-domain structure.

\noindent \textbf{Skill Breath} \quad
We further characterize tasks by the number of distinct skills they require. While this measure does not capture all dimensions of procedural complexity, it provides a proxy for the breadth of capabilities involved.
In \autoref{fig:no-complexity} (bottom), we report the average number of fine-grained skills per example for each benchmark. The resulting distribution reveals substantial heterogeneity: while 27.0\% of tasks require only a single skill, a nontrivial fraction (32.6\%) involve four or more distinct skills, reflecting substantially more complex procedures, e.g., the \texttt{TheAgentCompany} task in \autoref{fig:taxonomy}.

\section{How Autonomous Can Agents Be?}
\label{sec:3:autonomy}

Within the set of work currently represented in agent benchmarks, a central question is how autonomously they can reliably act at work. 
This question is often framed as a dichotomy between automation and augmentation. We argue, however, that these are not mutually exclusive modes. Whether an agent automates or augments depends on the scope of the task: a system may fully automate a narrowly scoped task (e.g., implementing bubble sort) while only augmenting within a more complex workflow (e.g., preparing a lecture on algorithms).
We therefore treat autonomy as a spectrum and examine agent autonomy levels as a function of task complexity.
In this section, we first quantify agent autonomy (\S\ref{sec:4.1:measure-autonomy}), and then provide practical guidance on selecting appropriate autonomy levels in agent deployment (\S\ref{sec:4.2:result-analysis}).

\subsection{Grounding Agent Autonomy in Task Complexity}
\label{sec:4.1:measure-autonomy}

Automating tasks can reduce human effort during execution, but raises the need for oversight to ensure correctness and control \citep{bainbridge1983ironies,parasuraman1997humans}. A key prerequisite for effective deployment is therefore understanding the autonomy level at which an agent operates.
To make this notion precise, we introduce operational definitions of \textit{task complexity} and \textit{agent autonomy}, which allow us to quantify autonomy levels and compare them across tasks and domains.

\begin{definition}
\label{def:complexity}
\textbf{Task Complexity}.
    The number and organization of distinct skills and procedural steps required to complete a task.
\vspace{-2mm}
\end{definition} 

Following \citet{wood1986task}, task complexity can be decomposed into component complexity (the number of distinct steps involved), coordinative complexity (i.e., interdependence among task elements), and dynamic complexity (i.e., uncertainty or environmental change). 
In this work, our definition corresponds primarily to component complexity, as the other two dimensions are difficult to estimate reliably from benchmark task descriptions.

\begin{wrapfigure}[11]{r}{0.28\textwidth}
\vspace{-2mm}
\small
\includegraphics[width=0.27\textwidth]{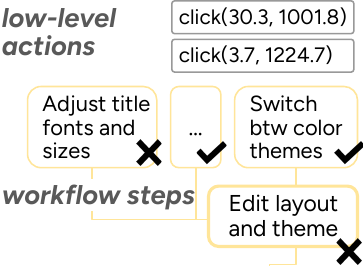}
\vspace{-1mm}
\caption{Exemplar workflow induced from agent low-level action trajectory.}
\label{fig:workflow-exemplar}
\end{wrapfigure} 
\noindent \textbf{Measuring Task Complexity} \quad
To obtain tasks with varied complexity, we adopt the workflow induction procedure from \citet{wang2025ai}, which segments any agent trajectory$\tau$ with low-level actions (e.g., \texttt{click}) into a hierarchical workflow $w$ with goal-directed steps at increasing levels of granularity (\autoref{fig:workflow-exemplar}). We quantify the complexity levels of these agent workflow steps (each representing a task of varied granularity), and use them to derive the autonomy levels above.

We approximate task complexity by the number of the most granular workflow steps $G = \left\{ v \in V \;\middle|\; \min_{\ell \in L}\mathrm{dist}(v,\ell)=0 \right\}$, as this level yields the most consistent representation of task structure, across workflows induced from heterogeneous trajectories from different agents and tasks.
$$\mathrm{complexity}(v) = \left|\, G \cap \mathrm{Desc}(v) \,\right|$$
where $\mathrm{Desc}(v)=\{u\in V \mid v \rightsquigarrow u\}$ is the set of descendants of $v$. 

Traditional behavioral measures of task complexity often conflate intrinsic task demands with variation in human ability  \citep{hackman1969toward}. For instance, using human completion time as a complexity indicator \citep{tamkinmccrory2025productivity,metr2025how} entangles task difficulty with variation in human capabilities, potentially biasing the resulting estimates.
Similarly, raw agent action counts depend heavily on the specifics of an agent's action space and execution strategy. In contrast, workflow steps abstract away these execution-level idiosyncrasies by collapsing repetitive low-level actions into semantically independent units that better reflect the underlying task structure.

To validate our approximation, we need to verify that: tasks at level-$k+1$ are strictly more complex\footnote{By ``more complex'', we mean higher perceived procedural complexity, as reflected in the number of steps a human would anticipate when mentally planning the task without execution.} than tasks at level-$k$; correspondingly, we can conclude that tasks within the same level exhibit comparable complexity. 
To validate this assumption, we perform pairwise comparisons between tasks at adjacent levels by prompting an LM to judge, based on task descriptions, whether a level-$k+1$ task is more complex than the level-$k$ task. We sample 10 pairs across all complexity levels from all benchmarks and agent trajectories, and find that this relative granularity criterion is satisfied in 82.6\% of comparisons, providing empirical support for the appropriateness of our workflow-based complexity approximation.

\begin{figure*}[t!]
\centering
    \includegraphics[width=\textwidth]{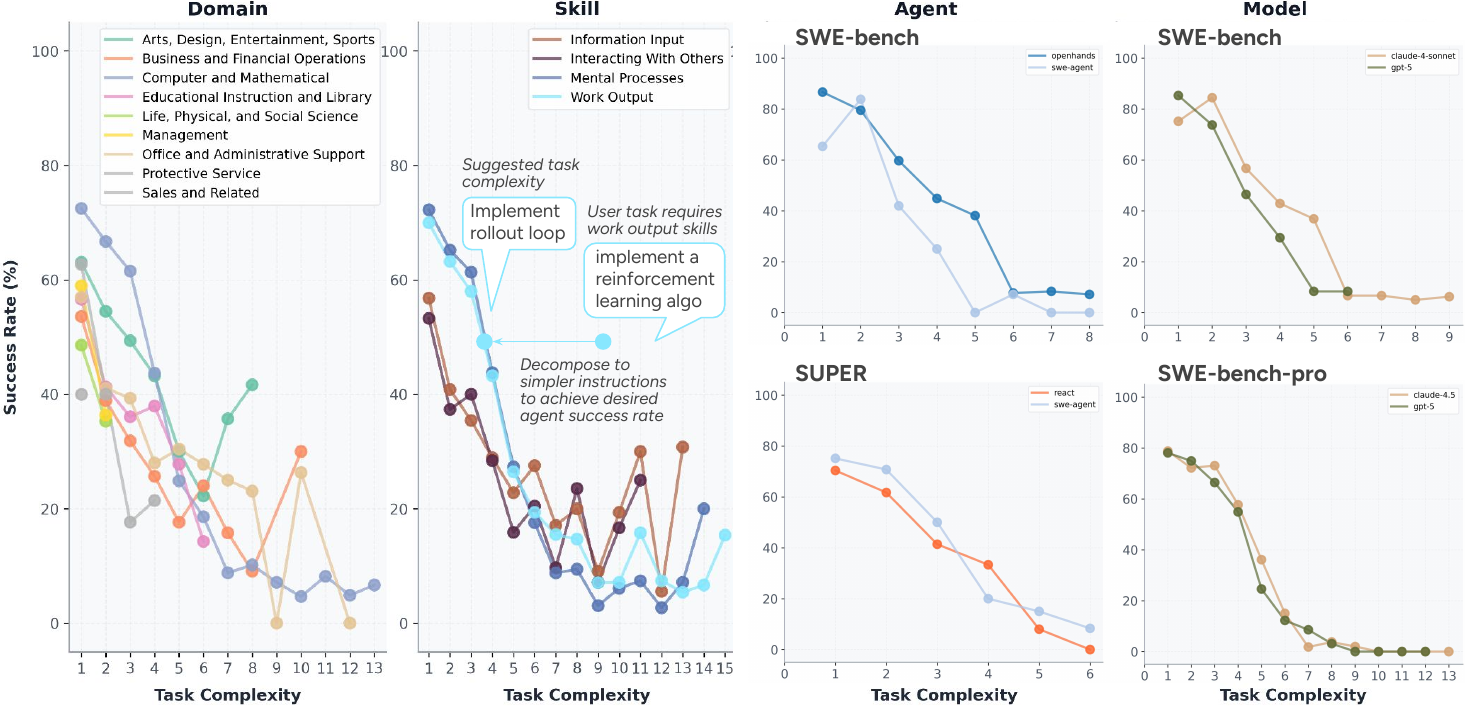}
\caption{Agent autonomy levels by (left) work domains and skills, and (right) agent frameworks and LM backbones.}
\label{fig:autonomy-exemplar}
\vspace{-2mm}
\end{figure*}

\begin{definition}
\label{def:autonomy}
\textbf{Autonomy}.
  The extent to which an agent system can complete a task by perceive its environment, make decisions, and take actions without direct human intervention. 
\vspace{-2mm}
\end{definition} 

Operationally, we quantify agent \textit{autonomy} as the maximum task complexity that an agent can complete end-to-end above a predefined success rate threshold with statistical confidence. This definition captures autonomy as a capability boundary: how complex a task an agent can reliably handle without human assistance.

\noindent \textbf{Measuring Agent Autonomy} \quad
Given tasks grouped by complexity level, we compute an agent's success rate at each level $k$ as $SR(k) = \frac{1}{|V^k|} \sum_{v \in V^k} \mathrm{status}(v)$, where $V^k = \{ v: \mathrm{complexity}(v) = k \}$ is the set of tasks at complexity level $k$, $\mathrm{status}(v) \in \{0, 1\}$ means step failure and success from the workflow induction procedure (\autoref{fig:autonomy-exemplar}).
Given a target success-rate threshold $H$, we define the \textit{autonomy level} of an agent as the highest task complexity at which the agent succeeds over:
\[
\mathrm{Autonomy} = \max \left\{ k \;\middle|\; SR(k) \ge H \right\}.
\]
This definition characterizes agent autonomy as a capability boundary, capturing the most complex class of tasks the agent can complete end-to-end with acceptable reliability.
Further, this characterization enables users to select appropriate tasks for automation and to calibrate the level of human oversight required for safe and effective deployment.

\subsection{Calibrating Agent Autonomy for Real-World Tasks}
\label{sec:4.2:result-analysis}
\vspace{-1mm}

We collect available agent trajectories for all benchmarks in our study (\S\ref{sec:2:bridge-agent-work}), and use them to analyze agent autonomy upon task complexity measures.
\autoref{fig:autonomy-exemplar} presents a breakdown of autonomy measures by (left) domain and skill requirements across all agent trajectories, and (right) agent framework and backbone LMs for benchmarks that allow fair comparisons.

\noindent $\bullet$ \textbf{Work Expertise}: 
Except for computer, business, and office-related domains, most domains show little to no coverage at higher task complexities, i.e., most tasks have complexity no greater than 6. 
Even in domains where agents perform best, most notably software engineering, success rates drop sharply as task complexity increases, despite the heavy concentration of tasks in these areas. In contrast, agents show greater autonomy for design tasks.

Across the four major skills, agents perform best on self-contained activities such as mental processes and work output, but struggle identifying and retrieving information (information input) or coordinating with others (interacting with others), particularly for simpler tasks.

\noindent $\bullet$ \textbf{Agent Selection}: 
A core challenge in comparing agent autonomy is that different benchmarks often evaluate different agent frameworks and LM backbones, making controlled comparisons difficult.
On the few coding-focused benchmarks where such comparisons are possible (e.g., SWE-bench), our analysis signals framework (OpenHands over SWE-agent) and LM (Claude over GPT) advantages, particularly for tasks of medium complexity. 
However, these trends may not be consistent across complexity regimes, motivating broader release of agent trajectories to enable more systematic and reproducible autonomy analysis.

\noindent \textbf{Choose the Right Autonomy Level for Your Agent} \quad
\vspace{-1mm}
Given a user task with desired performance requirements, we provide a principled strategy to \textit{decide which autonomy level the agent operates at}.
As demonstrated in \autoref{fig:autonomy-exemplar}, we consider a task (``implement a reinforcement algorithm'') with a target performance threshold $H$ (e.g., 80\%). 
Using our task-to-work mapping (\S\ref{sec:2:bridge-agent-work}), the system identifies relevant domains and skills to consult the relevant autonomy curves.

In this case, our system suggests that end-to-end execution at the original task complexity falls below the target $H$, thus recommends decomposing the task (e.g., using an LM) into simpler subtasks (e.g., ``implement rollout loop'') to assign to the agent for higher success. This example shows how autonomy curves translate abstract performance metrics into actionable decisions for human users.
Meanwhile, for agent developers, this curve exposes agent capability boundaries, thus facilitating more targeted development such as crafting tasks at this complexity for training.

\section{Discussion}
\label{sec:5:discussion}

\subsection{Related Work}
\noindent \textbf{AI's Impact on Work} \quad
A central motivation for developing AI agents is their potential to boost productivity in human work. Prior studies have examined AI's impact on employment \citep{brynjolfsson2025canaries}, productivity \citep{tamkinmccrory2025productivity}, and broader economic outcomes \citep{eloundou2023gpts,handa2025economic}.
In parallel, researchers have sought to build AI systems explicitly for work, including benchmarking agents at work \citep{patwardhan2025gdpval,xu2024theagentcompany}, studying workflows to integrate AI into work \citep{wang2025ai}, and surveys of industry practices and needs \citep{shome2025johnny,pan2025measuring,anthropic2025how,shao2025future}.
Despite this progress, we still lack a unified framework for characterizing agent utility in real-world work, which this work aims to address.

\noindent \textbf{Profiling AI Agents} \quad
Profiling AI systems offers practical guidance for both developers and users. Prior works mostly profile LLMs by personality or stylistic traits \citep{dunlap2024vibecheck}, studies of AI agents focus on specific domains (e.g., engineering \citep{metr2025measuring}) or surface-level architectural depiction \citep{casper2025ai}.
Despite existing conceptual frameworks \citep{feng2025levels}, agent autonomy remains vaguely defined and measured. We thus propose a quantifiable autonomy measure with practical usage guidance.

\noindent \textbf{Agent Benchmark Design} \quad
Benchmarks play a central role in shaping agent development, yet prior efforts often bias to easily specified or verifiable tasks \citep{jimenez2024swebench}. 
Despite the few benchmarks targeting human work \citep{xu2024theagentcompany,patwardhan2025gdpval}, we lack a systematic understanding of how benchmarks align with the distribution of real work, which we answer in this work.

\subsection{Principles to Benchmark Agents for Work}
\label{sec:4.5:principles}
Based on our analysis, we distill three practical principles to guide the design of benchmarks that better reflect real work.

\noindent \textbf{Domain and Skill Coverage} \quad
Rather than further focusing on already well-covered domains, benchmarks should either target underrepresented yet highly digitized domains, such as Management and Legal, which also account for a disproportionate share of capital; or aim for broad coverage across domains (as reported in \autoref{tab:benchmark-overview}). On the skill side, benchmarks should move beyond overemphasized granular skills (e.g., Getting Information, Working with Computers) and strive for a more balanced distribution across skill categories (e.g., Interacting with Others).
\autoref{fig:emp-domain} and \autoref{fig:skill-leaf} can act as a guide to identify these domains, and the supplementary website will be updated with new benchmarks as they become available.

\noindent \textbf{Ensure Realism and Complexity} \quad
Our analysis in \autoref{fig:no-complexity} indicates that many automatically synthesized benchmarks (e.g., ColBench) exhibit low domain and procedural complexity, capturing only simplified fragments of real work. In contrast, human-annotated tasks (e.g., in GDPval or TheAgentCompany) often involve more diverse domains and skills. 
While human annotation remains the gold standard, when synthesis is required for scalability, task generation should be grounded in realistic domain and skill compositions rather than abstract templates.

\noindent \textbf{Granular Evaluation} \quad
Our analysis in \autoref{fig:autonomy-exemplar} shows that singular end-task evaluations obscure differences in agents across task complexity. Although benchmarks with intermediate checkpoints \citep{xu2024theagentcompany} offer more informative assessments, it is costly to scale. As a more practical alternative, human demonstrations can be used to induce workflows \citep{wang2025ai}, enabling granular evaluation via automatically produced intermediate checkpoints.

\section{Conclusion}
This work examined whether and how current AI agent benchmarks reflect real-world work. By situating agent benchmarks within taxonomies of human work domains and skills, we revealed systematic mismatches between where agent development is concentrated and where real-world labor and economic value are distributed. 
We further measure agent autonomy across work, and translate these insights into practical guidance for different stakeholders:
enabling agent benchmark designers to assess gaps in work coverage, agent builders to identify areas of improvement, and agent users to select appropriate agents and autonomy levels for their specific work.
We hope this work motivates agent benchmarking and development efforts to more faithfully capture the diversity, complexity, and societal importance of real-world work.

\section*{Acknowledgments}
We would like to thank Yijia Shao, Pranjal Aggarwal, Ruiqi Zhong, and many members of the Language Technologies Institute for their helpful discussions and insightful feedback on the project. 
Zora Zhiruo Wang is supported by Google PhD Fellowship.


\newpage
\bibliography{main}

\newpage
\appendix

\section{Discussions}
\label{app:a:work-landscape}

\subsection{Representing Domains using O*NET}
\label{app:2.2:onet-domains}
\citet{handa2025economic} directly leverages O*NET data yet relies on LMs to synthesize a task hierarchy with limited human supervision and real work representativeness.
\citet{patwardhan2025gdpval} adopts industry sector annotations from the U.S. Bureau of Labor Statistics.
We directly adopt the official job family structure from the ONET database to define work domains, aligning our taxonomy with established occupational classifications.

Regarding the domain taxonomy, \autoref{fig:emp-domain} illustrates all top-level job families. Please find the full taxonomy at \url{https://github.com/zorazrw/ai4work-resources/blob/main/real_work/taxonomy/taxonomy_domain.json}.

\subsection{Representing Skills using O*NET}
\label{app:2.1:onet-skills}

We refer to ``skills'' based annotation, where a skill refers to a particular variety of workflow that an agent must be good at following standards in academia \citep{wang2025inducing,zheng2025skillweaver} and industry \citep{anthropic2025agentskills}.
To avoid confusion, we note that in addition to the \textit{Work Activities} annotation we adopted in \S\ref{sec:2:bridge-agent-work}, O*NET also provides annotations of ``skills'', including \textit{Skills} and \textit{Technical Skills}.
Nonetheless, these annotations are more conceptual (e.g., ``active learning''), focusing more on the developed capacities that facilitate learning or the more rapid acquisition of knowledge, and rarely grounded in tangible contexts of real human work (e.g., \textit{writing} skill in \textit{financial} and \textit{legal} domains).
To bridge this gap, we use the terminology from the agent development literature, but the data from ``work activities''.

\begin{figure*}[t!]
\centering
    \includegraphics[width=\textwidth]{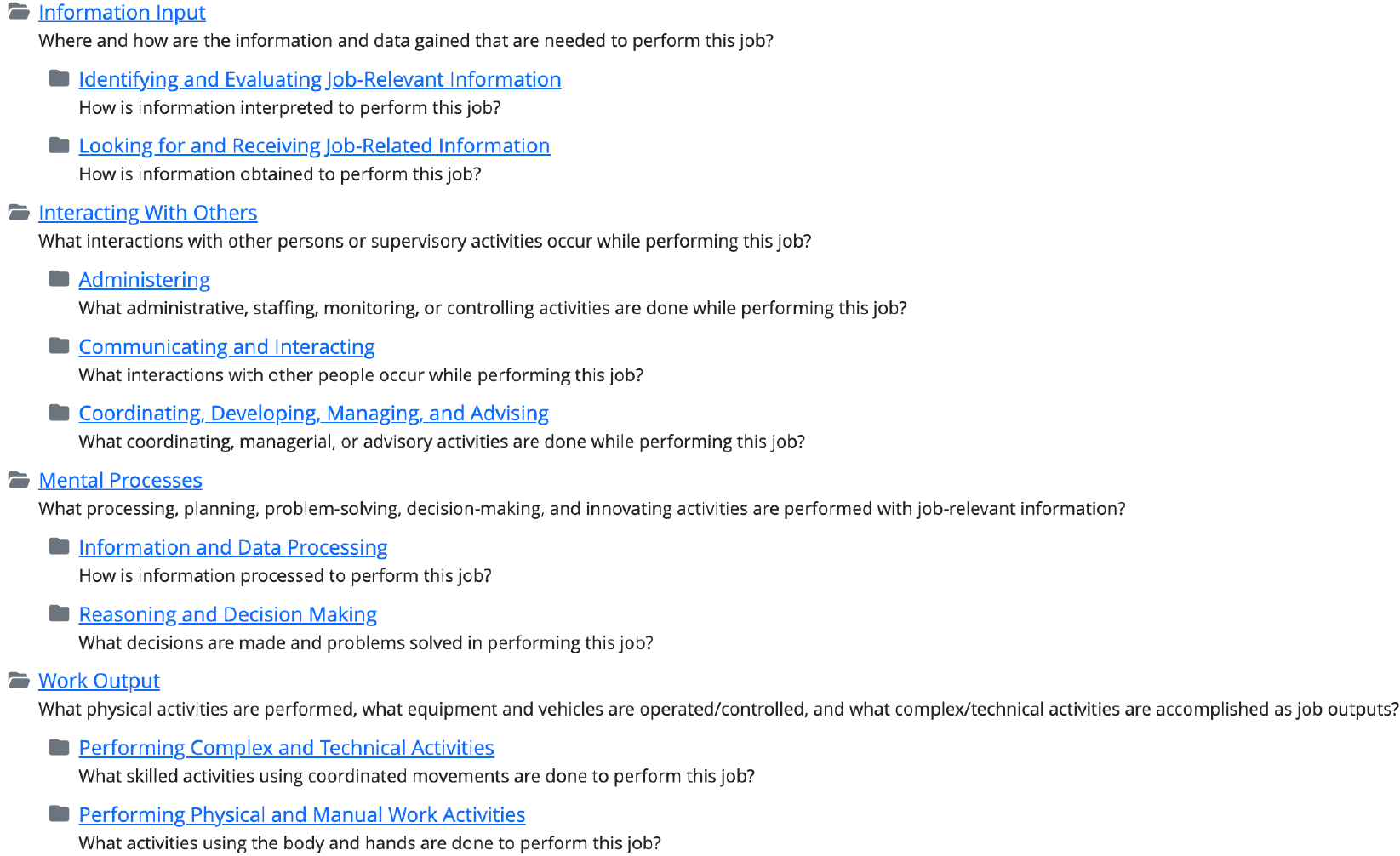}
\caption{Work activities annotation provided by O*NET.}
\label{fig:skill-wa}
\vspace{-2mm}
\end{figure*}

\autoref{fig:skill-wa} shows the first two layers work activities annotation provided in O*NET, which we adopt in our skill taxonomy. Due to space limitations, please find the full skill taxonomy at: \url{https://github.com/zorazrw/ai4work-resources/blob/main/real_work/taxonomy/taxonomy_skill.json}.

\subsection{Agentic Benchmark Selection}
\label{app:b:agent-dev-details}

\paragraph{The Focus on \textit{Agentic} Benchmarks}

Agentic benchmarks differ from traditional AI benchmarks in how tasks are structured and evaluated (\citep{zhu2025establishing,kim2025towards}). Traditional benchmarks typically assess performance on single-step tasks with fixed procedures (e.g., classification or short-form generation), where success is measured by matching a predefined output. In contrast, agentic benchmarks evaluate systems on multi-step tasks that require planning, tool use, and adaptive decision-making to produce an end-to-end outcome. Performance is therefore assessed based on whether the agent successfully completes the full task rather than a single intermediate prediction.

Because single-step benchmarks optimize for isolated operations, they often produce specialized tools that operate in narrowly scoped settings and offer limited insight into how AI systems interact with complex, real-world workflows. Agentic systems, by contrast, are designed to perform multi-step tasks over longer horizons and under changing conditions. Agent benchmarks therefore more directly probe the capabilities needed for deploying AI systems that can complete tasks independently or support humans across extended workflows—an important open question for real-world use.

At the same time, traditional single-step benchmarks remain compatible with our autonomy-level analysis. They typically correspond to low-autonomy (level-1) tasks that serve as building blocks for more complex workflows. Many work domains that appear underrepresented in current agent benchmarks already have strong coverage of such low-level subtasks in existing benchmarks, but these resources are rarely integrated into agent evaluation. We therefore advocate for tighter integration between traditional task benchmarks and agent-centric evaluations to support agents that generalize across a broader range of work.

\paragraph{Benchmarks Related to Physical Work}
To improve coverage of physical work, we surveyed existing robotics benchmarks and incorporated two representative datasets that include natural-language task descriptions and multi-step workflows. However, several limitations remain. Many robotics benchmarks primarily focus on low-level motor primitives (e.g., grasping, pushing, navigation), which do not map cleanly to real-world occupational tasks at the level of abstraction considered in this paper. Among benchmarks featuring more complex tasks, a substantial portion rely on visual goal specifications rather than natural-language instructions, making consistent taxonomy-based mapping difficult. Even in benchmarks that provide natural-language task descriptions, instructions are often highly templated and relatively simple (e.g., ``pick up trash'' or ``move object to location''), limiting their representativeness of realistic physical work contexts.

As a result, although the inclusion of these datasets improves coverage of physical tasks, the overall representation of real-world physical work remains limited. We therefore interpret our findings regarding physical domains with appropriate caution and view the development of richer, work-representative robotics benchmarks as an important direction for future research.

\subsection{Distinguishing Digital and Physical Work}
As introduced in \S\ref{sec:2.1:work-landscape}, we use an LLM to determine whether a task requirement requires digital or physical work.
More specifically, we adopt \texttt{claude-sonnet-3.5} for efficient annotation of this relatively simple task. We fill in the task requirement using the following prompt:

\begin{tcolorbox}[colback=gray!10, colframe=gray!60, coltitle=black, title=Prompt for Evaluating Goal-Action Consistency, fonttitle=\bfseries]
You are given an occupational task description, and your task is to classify whether completing this task primarily requires digital work or physical work.

Task: \{task-description\}

Return: DIGITAL or PHYSICAL. Provide a one-sentence justification.
\end{tcolorbox}

\section{Benchmark Mapping}
\label{app:b:mapping-validation}

\subsection{Validating LLM Mapping Results}

To verify the LM-based mapping, we randomly sample 20 examples from each benchmark to ensure coverage across all benchmarks, rather than sampling from a single pooled set.
We ask three graduate student annotators from the authors' institution, who are familiar with the O*NET taxonomy and the study's annotation guidelines, to grade the mapped work paths following the rubrics in \autoref{tab:annotation-rubrics}.
Annotators were instructed to review each task description and judge whether the mapped work path (domain and skill) appropriately captured the core intent of the task.
In our manually verified sample, the LM's predicted domain and skill paths align with the human judgment for 92\% and 93\% of the time, respectively. Remaining disagreements are relatively minor and primarily arise from under-specified natural language task descriptions.
For example, a task like ``build a music recommendation system'' could map to the \textit{Media} domain given its content focus, or to the \textit{Computer and Mathematical} domain given that implementation typically involves programming. Without explicit information about the intended execution, human and LM annotators sometimes infer the most appropriate domain differently.
Empirically, these disagreements occur primarily in the ColBench and GAIA benchmarks, likely because their task formulations are less explicitly grounded in real-world work contexts.

\begin{table}[t!]
\centering
\small
\resizebox{0.99\linewidth}{!}{
\begin{tabular}{l|rcc|cc|cc}
\toprule
\multicolumn{1}{c}{\multirow{2}{*}{\bf Benchmark}} & \multicolumn{3}{c}{\bf Size} & \multicolumn{2}{c}{\bf Domain Coverage (\%)}  & \multicolumn{2}{c}{\bf Skill Coverage (\%)} \\
{} & \multicolumn{1}{c}{total} & {avg} & {95\% CI} & {avg} & {95\% CI} & {avg} & {95\% CI} \\
\midrule
{TheAgentCompany} & {175} & {175} & {[175.0,175.0]} & {62.1} & {[62.1,62.1]} & {60.7} & {[60.7,60.7]} \\
{GDPval} & {220} & {220} & {[220.0,220.0]} & {77.0} & {[77.0,77.0]} & {82.7} & {[82.7,82.7]} \\
{Remote Labor Index} & {9} & {9} & {[9.0,9.0]} & {27.8} & {[27.8,27.8]} & {35.7} & {[35.7,35.7]} \\
{WorkArena} & {300} & {99.6} & {[97.2,102.0]} & {82.5} & {[81.7,83.4]} & {69.6} & {[68.1,71.2]} \\
{OfficeBench} & {300} & {186.0} & {[183.5,188.5]} & {73.4} & {[72.5,74.3]} & {82.3} & {[81.6,82.9]} \\
{CRMArena} & {300} & {52.2} & {[50.2,54.2]} & {74.9} & {[73.3,76.5]} & {83.4} & {[82.3,84.6]} \\
{EnterpriseBench} & {340} & {324.0} & {[321.9,326.1]} & {70.5} & {[70.2,70.8]} & {55.4} & {[55.1,55.8]} \\
{GitTaskBench} & {54} & {53.5} & {[53.0,53.9]} & {61.1} & {[60.8,61.5]} & {85.8} & {[85.7,85.9]} \\
\midrule
{OSWorld} & {390} & {370.5} & {[367.9,373.1]} & {72.8} & {[72.5,73.1]} & {79.5} & {[79.2,79.9]} \\
{WebVoyager} & {300} & {110.0} & {[106.9,113.0]} & {72.2} & {[70.9,73.5]} & {67.4} & {[66.1,68.6]} \\
{WebArena} & {300} & {273.3} & {[270.8,275.8]} & {76.0} & {[75.8,76.3]} & {88.0} & {[87.5,88.5]} \\
{Mind2Web} & {300} & {182.2} & {[177.3,187.2]} & {644.4} & {[63.1,65.8]} & {65.2} & {[63.8,66.7]} \\
{WebShop} & {300} & {14.4} & {[13.6,15.2]} & {89.5} & {[88.6,90.4]} & {93.8} & {[93.2,94.4]} \\
{VisualWebArena} & {300} & {88.9} & {[85.7,92.0]} & {70.5} & {[69.0,71.9]} & {75.5} & {[74.1,76.9]} \\
{WebLINX} & {99} & {98.6} & {[98.1,99.1]} & {68.9} & {[68.7,69.0]} & {65.8} & {[65.7,66.0]} \\
{AppWorld} & {300} & {200.1} & {[197.5,202.6]} & {79.6} & {[79.2,80.1]} & {74.5} & {[73.5,75.5]} \\
{AssistantBench} & {214} & {166.9} & {[164.9,169.0]} & {87.7} & {[87.5,88.0]} & {87.8} & {[87.1,88.5]} \\
{SPA-Bench} & {300} & {193.1} & {[189.6,196.6]} & {71.3} & {[70.5,72.1]} & {72.8} & {[71.8,73.9]} \\
{MMInA} & {300} & {63.6} & {[60.9,66.2]} & {79.8} & {[78.3,81.2]} & {84.4} & {[83.3,85.6]} \\
{WebChoreArena} & {300} & {169.0} & {[166.3,171.7]} & {78.6} & {[78.0,79.3]} & {77.4} & {[76.2,78.5]} \\
\midrule
{GAIA} & {300} & {241.7} & {[237.1,246.2]} & {62.8} & {[61.9,63.6]} & {60.1} & {[58.9,61.3]} \\
{TravelPlanner} & {300} & {8.0} & {[7.6,8.4]} & {95.4} & {[94.9,95.9]} & {97.3} & {[97.0,97.6]} \\
{DeepPlanning} & {120} & {15.0} & {[14.0,16.0]} & {90.4} & {[89.5,91.3]} & {97.8} & {[97.4,98.3]} \\
\midrule
{SWE-bench} & {300} & {89.5} & {[86.7,92.4]} & {80.0} & {[79.0,81.0]} & {69.6} & {[68.1,71.1]} \\
{TerminalBench} & {232} & {232.0} & {[232.0,232.0]} & {76.4} & {[76.4,76.4]} & {70.4} & {[70.4,70.4]} \\
{ColBench} & {860} & {815.3} & {[808.8,821.7]} & {68.1} & {[67.8,68.3]} & {67.7} & {[67.4,68.1]} \\
{SWE-Lancer} & {198} & {43.8} & {[41.9,45.7]} & {79.1} & {[77.7,80.4]} & {77.8} & {[76.4,79.1]} \\
{SWE-Bench MM} & {300} & {75.3} & {[73.3,77.2]} & {84.7} & {[83.8,85.6]} & {69.4} & {[67.7,71.1]} \\
{SWE-Bench Pro} & {300} & {150.8} & {[147.4,154.2]} & {78.6} & {[77.9,79.4]} & {72.5} & {[71.2,73.9]} \\
{MLE-Bench} & {82} & {81.9} & {[81.6,82.1]} & {73.7} & {[73.6,73.8]} & {76.5} & {[76.4,76.5]} \\
{SWT-Bench} & {300} & {16.3} & {[15.3,17.2]} & {90.1} & {[89.2,90.9]} & {94.0} & {[93.5,94.5]} \\
\midrule
{DiscoveryBench} & {300} & {103.7} & {[100.7,106.7]} & {84.1} & {[83.4,84.8]} & {84.5} & {[83.3,85.8]} \\
{ScienceAgentBench} & {102} & {46.4} & {[45.0,47.7]} & {87.8} & {[87.0,88.6]} & {83.2} & {[82.1,84.2]} \\
{CORE-Bench} & {45} & {42.2} & {[41.3,43.0]} & {61.3} & {[60.4,62.2]} & {62.8} & {[61.9,63.6]} \\
{SciCode} & {80} & {25.6} & {[24.5,27.4]} & {87.9} & {[86.9,89.0]} & {92.1} & {[91.4,92.7]} \\
{MLGym} & {19} & {17.8} & {[17.5,18.0]} & {83.6} & {[83.2,84.1]} & {75.0} & {[74.3,75.8]} \\
{DiscoveryWorld} & {59} & {58.3} & {[57.8,58.8]} & {83.0} & {[82.7,83.4]} & {80.3} & {[80.1,80.5]} \\
{LabBench} & {300} & {96.5} & {[93.6,99.5]} & {72.6} & {[71.4,73.7]} & {76.2} & {[74.9,77.5]} \\
{SUPER} & {300} & {34.5} & {[33.1,35.8]} & {87.3} & {[86.4,88.2]} & {83.4} & {[82.3,84.5]} \\
\midrule
{Sotopia} & {300} & {207.9} & {[202.9,212.8]} & {65.3} & {[64.2,66.5]} & {73.0} & {[72.2,73.8]} \\
{STSS} & {40} & {39.8} & {[39.6,40.0]} & {51.8} & {[51.5,52.1]} & {62.2} & {[61.9,62.4]} \\
\midrule
{Behavioral-1K} & {$~~~~$50} & {49.6} & {[49.2,49.9]} & {23.7} & {[24.1,25.4]} & {57.5} & {[57.1,57.8]} \\
{FieldWorkArena} & {300} & {131.0} & {[128.2,133.7]} & {79.3} & {[78.5,80.1]} & {72.9} & {[71.6,74.3]} \\
\bottomrule
\end{tabular}
}
\caption{Permutation-based sensitivity analysis of the coverage-aware sampling procedure.
For each benchmark, we report the distribution (median and 95\% confidence interval over 500 permutations) of (i) the number of tasks selected and (ii) Chao1-estimated domain and skill coverage, before stopping.}
\label{tab:sampling-test}
\end{table}

\subsection{Permutation-Based Sensitivity Analysis}
To assess the robustness of our coverage-aware sampling procedure, we perform a permutation-based sensitivity analysis within each sampled benchmark subset. Specifically, for each benchmark, we randomly permute the order of the sampled tasks and replay our batch-wise stopping rule 500 times. For each run, we record (i) the number of tasks selected before the stopping criterion is met and (ii) the resulting domain and skill coverage at termination (\autoref{tab:sampling-test}).

Across benchmarks, the stopping point is stable: most runs terminate after processing a substantially small portion of the sampled subset, indicating that coverage saturation is consistently detected rather than driven by a particular task ordering. Moreover, the domain and skill coverage achieved at termination closely matches the maximum coverage attainable within the sampled subset, with narrow 95\% confidence intervals across permutations. These results suggest that our adaptive stopping rule is not strongly sensitive to task ordering and does not prematurely terminate within the sampled pool.

\begin{table}[t!]
\centering
\small
\begin{tabular}{p{2.8cm}p{7.cm}p{3.8cm}}
\toprule
\multicolumn{1}{c}{\bf Label} & \multicolumn{1}{c}{\bf When to Use} & \multicolumn{1}{c}{\bf Key Question} \\
\midrule
{All Correct} & 
All assigned locations are appropriate and no relevant locations are missing. The assignment set comprehensively and accurately captures the instruction's requirements.
& 
Are all assignments appropriate \textit{and} is the set complete? \\
\midrule
{All Wrong} & 
Multiple assigned locations are inappropriate and potentially need both addition/removal. The LLM has fundamentally misunderstood the task or assigned completely unrelated locations.
& 
Is every single assignment inappropriate or unrelated? \\
\midrule
{Missing Locations} & 
All current assignments are correct, but additional relevant locations should be included to fully capture the instruction's scope.
& 
Are the current assignments right, but incomplete? What's missing? \\
\midrule
{Extra Locations} & 
The assignment includes some correct locations but also contains extraneous assignments that are not required by the instruction.
& 
Which specific assignments are inappropriate and should be removed? \\
\bottomrule
\end{tabular}
\vspace{3mm}
\caption{Rubrics for manually validating LM-based mapping results.}
\label{tab:annotation-rubrics}
\end{table}

\section{Measuring Agent Autonomy}

We provide more concrete examples for understanding task complexity levels in featuring agent autonomy in \autoref{tab:task-by-complexity}.

\begin{table}[htbp]
\centering
\resizebox{\textwidth}{!}{%
\begin{tabular}{cl}
\toprule
\textbf{Level} & \multicolumn{1}{c}{\bf Exemplar Task} \\
\midrule
1 & {Navigate to the Plus section of Cambridge Dictionary to access the Contents menu for Image Quizzes.} \\
2 & {Navigating the Google Travel interface menu options.} \\
3 & {Locate the sparql command and verify rdflib Python module installation.} \\
4 & {Define functions to load and preprocess training, validation, and test datasets.} \\
5 & {Authenticate with phone system and retrieve contact relationship information.} \\
6 & {Retrieve API documentation to understand available endpoints and their functionality.} \\
7 & {Inspect JSON file structure and update script to use correct field names for data processing.} \\
8 & {Attempt to find fine-tuning instructions for the TransNormerLLM-385M model in the README file.} \\
9 & {Create test configuration, helper functions, and run initial 2A test suite for Raft implementation.} \\
10 & {Summarize the ModelAdmin class goal and pass the request object to inline class constructor.} \\
11 & {Train models using the improved training script with enhanced features and optimizations.} \\
12 & {Create and verify a filled PDF form with calculated tax credit values.} \\
13 & {Create and debug Python files for SQL generation in Django ORM's expression system.} \\
14 & {The goal is to implement a reinforcement learning algorithm.} \\
\bottomrule
\end{tabular}
}
\vspace{1mm}
\caption{Exemplar task instructions at varied task complexities.}
\label{tab:task-by-complexity}
\end{table}

\end{document}